\documentclass[journal]{IEEEtran}

\pdfoutput=1

\ifCLASSINFOpdf
  \usepackage[pdftex]{graphicx}

  \graphicspath{{./figures/}}

  \DeclareGraphicsExtensions{.pdf,.jpeg,.png}

\else

\fi

\usepackage[utf8]{inputenc}
\usepackage{amsmath}
\usepackage{hyperref}

\usepackage{algorithm}
\usepackage[noend]{algpseudocode}

\usepackage{amssymb}

\usepackage{color}
\usepackage[dvipsnames]{xcolor}
\usepackage{subcaption}

\usepackage{multirow}
\usepackage{booktabs} 

\begin{document}

\title{Using Centroidal Voronoi Tessellations \\to Scale Up the Multi-dimensional Archive of Phenotypic Elites Algorithm}


\author{Vassilis~Vassiliades, Konstantinos Chatzilygeroudis, and~Jean-Baptiste~Mouret
\thanks{All authors have the following affiliations: (1) Inria, Villers-lès-Nancy, F-54600, France; (2) CNRS, Loria, UMR 7503, Vandœuvre-lès-Nancy, F-54500, France; (3) Université de Lorraine, Loria, UMR 7503, Vandœuvre-lès-Nancy, F-54500, France; (e-mails: \{vassilis.vassiliades, konstantinos.chatzilygeroudis, jean-baptiste.mouret\}@inria.fr).}
\thanks{This work was supported by the European Research Council (ERC) under the European Union's Horizon 2020 research and innovation programme (Project: ResiBots, grant agreement No 637972).}
}

\maketitle

\begin{abstract}

The recently introduced Multi-dimensional Archive of Phenotypic Elites (MAP-Elites) is an evolutionary algorithm capable of producing a large archive of diverse, high-performing solutions in a single run. It works by discretizing a continuous feature space into unique regions according to the desired discretization per dimension. While simple, this algorithm has a main drawback: it cannot scale to high-dimensional feature spaces since the number of regions increase exponentially with the number of dimensions. In this paper, we address this limitation by introducing a simple extension of MAP-Elites that has a constant, pre-defined number of regions irrespective of the dimensionality of the feature space.
Our main insight is that methods from computational geometry could partition a high-dimensional space into well-spread geometric regions. In particular, our algorithm uses a centroidal Voronoi tessellation (CVT) to divide the feature space into a desired number of regions; it then places every generated individual in its closest region, replacing a less fit one if the region is already occupied. We demonstrate the effectiveness of the new ``CVT-MAP-Elites" algorithm in high-dimensional feature spaces through comparisons against MAP-Elites in maze navigation and hexapod locomotion tasks.

\end{abstract}


%
\IEEEpeerreviewmaketitle

\section{Introduction}

Evolution started from a common ancestor~\cite{theobald2010formal} and gave rise to the biodiversity we see today, which is estimated to amount to 1 trillion species~\cite{locey2016scaling}. Inspired by this observation, the field of evolutionary robotics has recently seen a shift from evolutionary algorithms (EAs) that aim to return a single, globally optimal solution, to algorithms that explicitly search for a very large number of diverse, high-performing individuals~\cite{lehman2011evolving, cully2013behavioral, clune2013modularity, mouret2015illuminating, pugh2015confronting, pugh2016quality, cully2016evolving, smith2016rapid}.

\begin{figure}
	\centering
	\includegraphics[width=0.7\columnwidth]{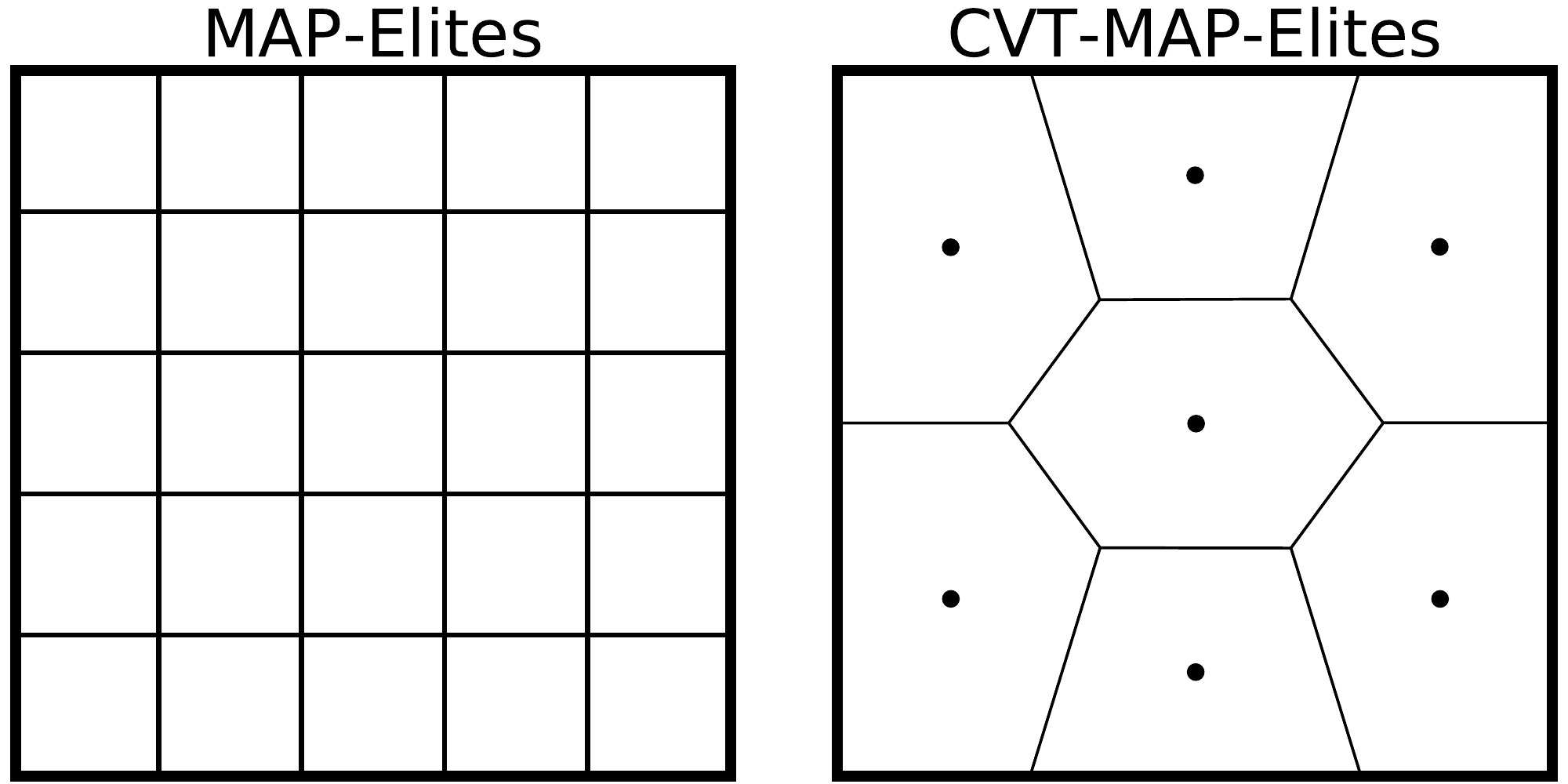}
	\caption{ MAP-Elites discretizes the feature space according to some pre-specified number of discretizations per dimension. This means that the number of niches grow exponentially with the number of additional dimensions or discretizations, thus, it cannot be used in high-dimensional feature spaces. In contrast, Centroidal Voronoi Tessellation (CVT) - MAP-Elites, uses a CVT~\cite{du1999centroidal} to partition the feature space into $k$ homogeneous geometric regions, where $k$ is the pre-specified number of niches. Here, MAP-Elites uses 5 discretizations per dimension, resulting in 25 niches, whereas, CVT-MAP-Elites uses 7 niches.
	}
	\label{fig_conceptual}
\end{figure}

\begin{figure*}[th]
	\centering
	\includegraphics[width=0.9\textwidth]{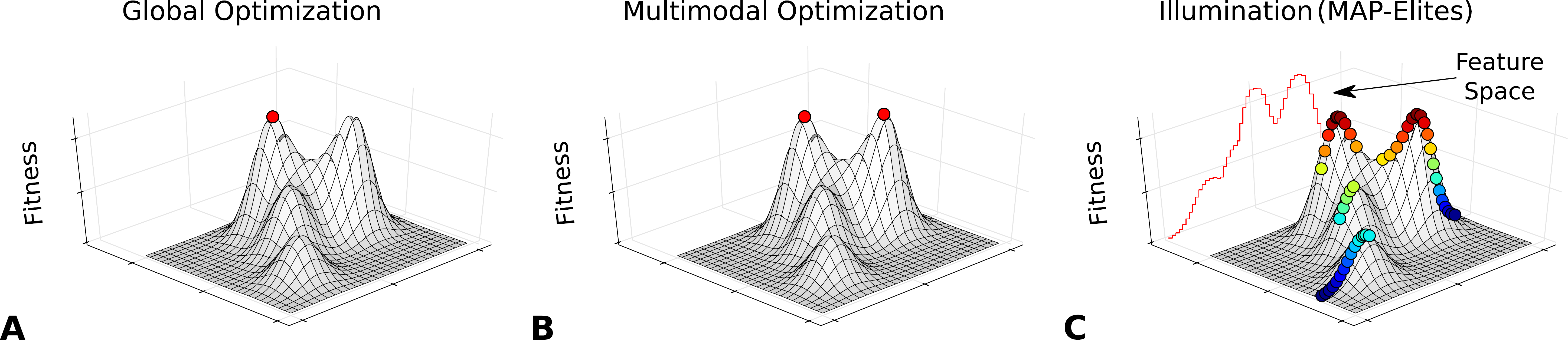}
	\caption{Algorithms for global optimization aim to find a single global optimum of the underlying parameter space in which they operate (\textbf{A}). Multimodal optimization algorithms, on the other hand, aim at finding multiple optima (\textbf{B}). Illumination algorithms, such as MAP-Elites (\textbf{C}) go one step further by aiming to discover significantly more solutions, each one being the \textit{elite} of some local neighborhood defined in some \textit{feature space} of interest. For finding the solutions of the function illustrated, we used (\textbf{A}) Covariance Matrix Adaptation Evolution Strategies~\cite{hansen2001completely}, (\textbf{B}) Restricted Tournament Selection~\cite{harik1995rts}, and (\textbf{C}) MAP-Elites~\cite{mouret2015illuminating}.}
	\label{fig_optimization_vs_illumination}
\end{figure*}

EAs have traditionally been used for \textit{optimization} purposes \cite{eiben_introduction_2015} with the aim of returning a single solution that corresponds to the global optimum of the underlying search space (Fig.~\ref{fig_optimization_vs_illumination}A). Various forms of diversity maintenance (niching) techniques have been designed to supply EAs both with the ability to avoid premature convergence to local optima and to perform \textit{multimodal optimization} \cite{mahfoud1995niching, harik1995rts, sareni1998fitness, das2011real, preuss2015multimodal}. In the latter case, the aim is to return multiple solutions that correspond to the peaks of the search space (Fig.~\ref{fig_optimization_vs_illumination}B).

A recent, alternative perspective in the field of evolutionary computation views EAs more as \textit{diversifiers}, rather than as optimizers~\cite{pugh2016quality}. Research on such EAs was initiated by the introduction of the Novelty Search (NS) algorithm~\cite{lehman2008exploiting, lehman2011abandoning}, which looks for individuals that are different from those previously encountered in some \textit{behavior space} or \textit{feature space} (we use both terms interchangeably).
The behavior space is defined by features of interest that describe the possible behaviors of individuals over their lifetime~\cite{doncieux2010behavioral, pugh2016quality}. For example, points in this space could be the final positions of a simulated robot in a 2D maze environment.
By rewarding novelty instead of fitness, NS promotes behavioral diversity and accumulates potential stepping stones for building more complex solutions~\cite{lenski2003evolutionary}. This algorithm relies on the intuition that it can be more beneficial to encourage exploration in the behavior space rather than the genotype space, which is confirmed experimentally for several domains~\cite{mouret2012encouraging, doncieux2014beyond}.

Yet, purely exploring the behavior space without considering the task performance is not practical in many cases. For example, we might not only be interested in finding controllers that make a robot reach different points in the environment, but also the fastest controller for each point. To address this issue, algorithms like NS with Local Competition~\cite{lehman2011evolving} and Multi-dimensional Archive of Phenotypic Elites (MAP-Elites)~\cite{mouret2015illuminating} explicitly maintain a large number of niches while optimizing them locally. For this reason they are called \textit{illumination} algorithms~\cite{mouret2015illuminating} or \textit{quality diversity} (QD) algorithms~\cite{pugh2015confronting, pugh2016quality}.
%
It is important to note that in multimodal optimization, the primary interest is \textit{optimization}, i.e., we are interested in maintaining diversity in order to find the peaks of the underlying genotype or phenotype space. On the contrary, the primary goal of illumination algorithms is \textit{not} optimization, but diversity~\cite{pugh2016quality}.

Whereas NS and its variants are governed by dynamics that continually push the population to unexplored regions in behavior space,
MAP-Elites~(Alg.~\ref{algo:map_elites}) uses a conceptually simpler approach: it discretizes the $d$-dimensional behavior space into unique bins (Fig.~\ref{fig_conceptual} left) by dividing each dimension into a specified number of ranges ($n_1, ... , n_d$); it then attempts to fill each of the $\prod_{i=1}^{d}n_i$ bins through a variation-selection loop with the corresponding genotype and its fitness, replacing a less fit one if it exists.
The final result is an \textit{archive} that contains the \textit{elite} solution of every niche in behavior space.
%
Being an illumination algorithm, MAP-Elites aims to return more information than multimodal optimization algorithms, i.e., solutions that are the best in their local region, but not necessarily ones where the fitness gradient is (near) zero (Fig.~\ref{fig_optimization_vs_illumination}C). The maximum number of solutions that can be returned (i.e., the number of niches) is controlled by the number of discretization intervals (i.e., $k=\prod_{i=1}^{d}n_i$) and is a \textit{user-defined input} to the algorithm. Put differently, the problem solved by MAP-Elites is ``find $k$ (e.g., 10,000) solutions that are as different and as high-performing as possible''.

MAP-Elites has been employed in many domains. For instance, it has been used to produce: behavioral repertoires that enable robots to adapt to damage in a matter of minutes~\cite{cully2015robots, tarapore2016different}, perform complex tasks~\cite{duarte2016evorbc}, or even adapt to damage while completing their tasks~\cite{chatzilygeroudis2016towards, chatzilygeroudis2016resetfree}; morphological designs for walking ``soft robots", as well as behaviors for a robotic arm~\cite{mouret2015illuminating}; neural networks that drive simulated robots through mazes~\cite{pugh2015confronting}; images that ``fool" deep neural networks~\cite{nguyen2015deep}; ``innovation engines" able to generate images that resemble natural objects~\cite{nguyen2016understanding}; and 3D-printable objects by leveraging feedback from neural networks trained on 2D images~\cite{lehman2016iccc}.

The \textit{grid-based} approach of MAP-Elites requires the user to only specify the number of discretization intervals for each dimension, making the algorithm conceptually simple and straightforward to implement. However, this approach suffers from the curse of dimensionality, since the number of bins increase exponentially with the number of feature dimensions. The increase in the number of niches results in reduced selective pressure, making the algorithm unable to cope with high-dimensional feature spaces even when memory is not a problem. For this reason, MAP-Elites has only been employed in settings with low-dimensional feature spaces (2 to 6 dimensions). However, scaling to high dimensions is a desirable property, as this would potentially allow MAP-Elites to be used with more expressive descriptors and create archives of better quality and diversity.

A way to address this limitation is by employing a method that maximally spreads a desired number of niches in feature spaces of arbitrary dimensionality. In this paper, we achieve this using a technique from computational geometry known as centroidal Voronoi tessellations (CVTs)~\cite{du1999centroidal}. In particular, the contribution of this paper is two-fold: (1) we introduce a new algorithm that we call ``CVT-MAP-Elites" (Fig.~\ref{fig_conceptual} right) and demonstrate its advantage over MAP-Elites in a maze navigation task and the simulated hexapod locomotion task of~\cite{cully2015robots}; (2) we propose a new methodology for assessing the quality of the archives produced by illumination algorithms.

\section{Centroidal Voronoi Tessellation MAP-Elites} \label{sec_CVT}

A Voronoi tessellation~\cite{aurenhammer2000voronoi} is a partitioning of a space into geometric regions based on distance to $k$ pre-specified points which are often called sites. Each region contains all the points that are closer to the corresponding site than to any other. If the sites are also the centroids of each region (and the space is bounded), then the Voronoi tessellation is the CVT~\cite{du1999centroidal} of the space (Fig.~\ref{fig_conceptual} right).
CVTs have found application in problems ranging from data compression to modeling the territorial behavior of animals~\cite{du1999centroidal}.

\begin{algorithm}
	\caption{CVT approximation (adapted from~\cite{ju2002probabilistic})}\label{algo:cvt}
  \begin{algorithmic}[1]
    \Procedure{CVT}{$k$}
    \State $\mathcal{C} \longleftarrow$ sample\_points($k$) \Comment{$k$ random centroids}
    \State $S \longleftarrow $ sample\_points($K$) \Comment{$K$ random samples}
    \For {i $= 0 \longrightarrow max\_iter$}
      \State $\mathcal{I} \longleftarrow $ get\_closest\_centroid\_indices($S,\mathcal{C}$)
      \State $\mathcal{C} \longleftarrow $ update\_centroids($\mathcal{I}$)
    \EndFor
    \State \textbf{return} centroids $\mathcal{C}$
    \EndProcedure
  \end{algorithmic}
\end{algorithm}

There exist various algorithms for constructing CVTs~\cite{hateley2015fast}. Lloyd's algorithm~\cite{lloyd1982least} is the most widely used in 2D spaces and consists of repeatedly constructing the Voronoi tessellation of the $k$ sites, computing the centroids of the resulting Voronoi regions and moving the sites to their corresponding centroids. However, explicitly constructing Voronoi tessellations in high-dimensional spaces involves complex algorithms~\cite{aurenhammer2000voronoi}. An alternative, simpler approach is to use Monte Carlo methods to obtain a close approximation to a CVT of the feature space~\cite{ju2002probabilistic}. In this work, we follow this approach and construct such an approximation using Alg.~\ref{algo:cvt} (adapted from~\cite{ju2002probabilistic}).
%
The algorithm first randomly initializes $k$ centroids (line 2) and generates $K$ random points ($K >> k$) (line 3) uniformly in the feature space according to the bounds of each dimension (e.g., the feature space could be defined in $[0,1]^d$).
The algorithm then alternates between assigning each point to the closest centroid and updating each centroid to be the mean of its corresponding points (lines 4-6). This procedure is analogous to using a clustering algorithm (such as $k$-means~\cite{macqueen1967kmeans}) to find $k$ clusters in a dataset that contains many well-spread points. Therefore, constructing a CVT can intuitively be seen as forcing the $k$ sites to be well-spread in the space of interest.

\begin{algorithm}
	\caption{MAP-Elites algorithm}\label{algo:map_elites}
	\begin{algorithmic}[1]
		\Procedure{MAP-Elites}{$[n_1,...,n_d]$}
    \State \colorbox{lightgray}{$(\mathcal{X}, \mathcal{P}) \longleftarrow$ create\_empty\_archive($[n_1,...,n_d]$)}
    \For {$i=1\to G$} \Comment{\emph{Initialization: $G$ random $\mathbf{x}$}}
      \State $\mathbf{x} = $ random\_solution()
      \State \textsc{add\_to\_archive}($\mathbf{x}, \mathcal{X}, \mathcal{P}$)
    \EndFor
    \For {$i=1\to I$} \Comment{\emph{Main loop, $I$ iterations}}
    	\State $\mathbf{x} = $ selection($\mathcal{X}$)
    	\State $\mathbf{x}' = $ variation($\mathbf{x}$)
      \State \textsc{add\_to\_archive}($\mathbf{x}', \mathcal{X}, \mathcal{P}$)
    \EndFor
		\State \textbf{return} archive ($\mathcal{X}, \mathcal{P}$)
		\EndProcedure

    \Procedure{add\_to\_archive}{$\mathbf{x}, \mathcal{X}, \mathcal{P}$}
    \State $(p,\mathbf{b}) \longleftarrow $ evaluate($\mathbf{x}$)
    \State \colorbox{lightgray}{$c \longleftarrow $ get\_cell\_index($\mathbf{b}$)}
    \If {$\mathcal{P}(c) = null$ or $\mathcal{P}(c) < p$}
    \State $\mathcal{P}(c) \longleftarrow p$, $\mathcal{X}(c) \longleftarrow \mathbf{x}$
    \EndIf
		\EndProcedure
	\end{algorithmic}
\end{algorithm}
\begin{algorithm}
	\caption{CVT-MAP-Elites algorithm}\label{algo:cvt_map_elites}
	\begin{algorithmic}[1]
		\Procedure{CVT-MAP-Elites}{$k$}
    \State \colorbox{lightgray}{$\mathcal{C} \longleftarrow$ \textsc{CVT}($k$) \Comment{Run CVT and get the centroids}}
    \State \colorbox{lightgray}{$(\mathcal{X}, \mathcal{P}) \longleftarrow$ create\_empty\_archive($k$)}
    \For {$i=1\to G$} \Comment{\emph{Initialization: $G$ random $\mathbf{x}$}}
      \State $\mathbf{x} = $ random\_solution()
      \State \textsc{add\_to\_archive}($\mathbf{x}, \mathcal{X}, \mathcal{P}$)
    \EndFor
    \For {$i=1\to I$} \Comment{\emph{Main loop, $I$ iterations}}
    	\State $\mathbf{x} = $ selection($\mathcal{X}$)
    	\State $\mathbf{x}' = $ variation($\mathbf{x}$)
      \State \textsc{add\_to\_archive}($\mathbf{x}', \mathcal{X}, \mathcal{P}$)
    \EndFor
		\State \textbf{return} archive ($\mathcal{X}, \mathcal{P}$)
		\EndProcedure

    \Procedure{add\_to\_archive}{$\mathbf{x}, \mathcal{X}, \mathcal{P}$}
    \State $(p,\mathbf{b}) \longleftarrow $ evaluate($\mathbf{x}$)
    \State \colorbox{lightgray}{$c \longleftarrow $ get\_index\_of\_closest\_centroid($\mathbf{b}, \mathcal{C}$)}
	    \If {$\mathcal{P}(c) = null$ or $\mathcal{P}(c) < p$}
	      \State $\mathcal{P}(c) \longleftarrow p$, $\mathcal{X}(c) \longleftarrow \mathbf{x}$
	    \EndIf
		\EndProcedure
	\end{algorithmic}
\end{algorithm}

CVT-MAP-Elites partitions the $d$-dimensional feature space into $k$ Voronoi regions and then attempts to fill each of the regions through a selection-variation loop. Algorithmically, it first obtains the coordinates of the $k$ centroids ($\mathcal{C}$; Alg.~\ref{algo:cvt_map_elites}, line 2) by constructing the CVT as described above (Alg.~\ref{algo:cvt}). It then creates an empty archive with capacity $k$ ($\mathcal{X}$ and $\mathcal{P}$ store the genotypes and performances, respectively). The algorithm then evaluates $G$ random parameter vectors ($\mathbf{x}$), simultaneously recording their performance ($p$) and feature \textit{descriptor} ($\mathbf{b}$), i.e., their location in feature space (Alg.~\ref{algo:cvt_map_elites}, 4-6). Next, it finds the index ($c$) of the centroid in $\mathcal{C}$ that is closest to $\mathbf{b}$ (Alg.~\ref{algo:cvt_map_elites}, line 14), which implicitly gives information about its Voronoi region. If the region is free, then the algorithm stores the parameter vector $\mathbf{x}$ in that region; if it is already occupied, then the algorithm compares the performance values and keeps only the best parameter vector (Alg.~\ref{algo:cvt_map_elites}, 15-16). Once this is done, CVT-MAP-Elites iterates a simple loop (Alg.~\ref{algo:cvt_map_elites}, 7-10): (1) randomly select one of the occupied regions to obtain the stored $\mathbf{x}$, (2) add some random variation on $\mathbf{x}$ to obtain $\mathbf{x}'$, (3) record the performance and feature descriptor, and (4) attempt to insert the new parameter in the corresponding region.

The differences between MAP-Elites and CVT-MAP-Elites are highlighted in Alg.~\ref{algo:map_elites} and Alg.~\ref{algo:cvt_map_elites}: MAP-Elites creates an empty archive based on the desired discretizations per dimension $[n_1,...,n_d]$ (Alg.~\ref{algo:map_elites}, line 2), whereas CVT-MAP-Elites first performs the CVT construction and then creates the empty archive of $k$ niches (Alg.~\ref{algo:cvt_map_elites}, line 2,3).
MAP-Elites calculates the index $c$ of a descriptor $\mathbf{b}$ in $O(1)$ time (Alg.~\ref{algo:map_elites}, line 13), whereas CVT-MAP-Elites needs to be equipped with a distance function for doing so (Alg.~\ref{algo:map_elites}, line 14); the time complexity for the nearest neighbor query could be $O(\log k)$ on average~\cite{bentley1975multidimensional, aurenhammer2000voronoi} or $O(k)$ in the worst case.

\section{Evaluation} \label{sec_evaluation}

To assess the scalability of CVT-MAP-Elites, we experiment with maze navigation and hexapod locomotion tasks. Both classes of tasks have been successful in testing the ability of illumination algorithms to explicitly generate archives that contain many diverse and high-performing solutions~\cite{cully2015robots, pugh2016quality}.

A few metrics have been used in the literature to evaluate illumination algorithms~\cite{mouret2015illuminating, pugh2015confronting, pugh2016quality}, many of which utilize the MAP-Elites grid. For example, ``coverage"~\cite{mouret2015illuminating} measures the expected number of cells an algorithm can fill using a specific descriptor, while ``quality diversity score"~\cite{pugh2015confronting, pugh2016quality} projects the descriptors to a certain feature space and calculates the sum of the fitness values stored in each cell.

These metrics, however, have two disadvantages that prevent us from using them. First, they are dependent on the behavior space and a particular discretization of MAP-Elites, whereas we would like to compare not only different EAs, but also spaces of different dimensionality (e.g., 2D vs 1000D). Second, they do not explicitly assess if the archives contain the ``right" diversity, where ``right" here is task-specific. For example, in our maze navigation experiments, we would like the resulting archives to contain controllers that take the robot from the starting position to the goal using different trajectories. In our hexapod locomotion experiments, we are interested in collecting diverse and high-quality solutions that would be useful even if the dynamics of the robot change due to some damage.

Since the aforementioned metrics cannot work in our case, we propose the following methodology for assessing the quality of the archives produced by each EA-descriptor pair $P_i$: in an analogy with supervised learning scenarios, training is done during the standard evolutionary phase, while testing for generalization is done during an \textit{evaluation} phase in settings \textit{not} experienced during evolution. Each evaluation setting slightly modifies the simulator in order to test whether different classes of behaviors are found by $P_i$, without however changing the fitness function (i.e., the way individuals are rewarded). For example, in the maze navigation experiments, an evaluation scenario $e$ would correspond to changing the environment by blocking certain paths of the maze in order to test whether controllers that achieve a particular trajectory are found by $P_i$. In the hexapod locomotion experiments, $e$ would correspond to changing the dynamics of the robot by removing a leg, in order to test whether $P_i$ has found controllers that perform well in spite of this damage. For each $P_i$, we generate multiple archives from independent evolutionary trials, in order to make statistical assessments, and use the following metrics:

\subsection{Expected best performance of an EA-descriptor pair in an evaluation scenario $e$}
In order to calculate the ``best performance'' of an archive returned by $P_i$, we exhaustively\footnote{In our experiments, in case an archive contains more than 10k solutions, we randomly select and evaluate 10k of them to reduce evaluation time.} evaluate all solutions contained in the archive by simulating the given evaluation scenario $e$, and return the fitness value of the fittest solution. To calculate the ``expected best performance'' of $P_i$ on $e$, we perform this procedure on the archives returned from all evolutionary runs and take the median value. For example, in the maze experiment where $e$ corresponds to allowing only a single open path towards the goal, this measure would capture whether $P_i$ was able to find at least one solution (controller) that makes the robot follow this path, even if this solution is the only one in the archive that achieves this. Intuitively, this metric asks: can $P_i$ find a solution for a test problem?

\subsection{Expected quality of an EA-descriptor pair}
The ``expected best performance'' metric is the extreme case of a more general metric that asks: how many solutions can $P_i$ find for a test problem? Since in our case ``solving a problem'' is not a discrete event, i.e., the fitness values are continuous variables, we can define this metric to be the probability that the fitness value $X$ drawn from the archives produced by $P_i$ is less than or equal (if we are minimizing; greater if we are maximizing) to a certain value $x$, where $x$ is task-specific. If we consider not just one value for $x$, but a range of values, we can generalize this metric by generating the cumulative distribution function (CDF), for a minimization problem, or the complementary CDF (CCDF), for a maximization problem. That is:
\begin{itemize}
  \item CDF of $P_i$: $F_{X \sim P_i}(x) = P(X \leq x)$
  \item CCDF of $P_i$: $\widetilde{F}_{X \sim P_i}(x) = P(X > x)$
\end{itemize}
We calculate these functions by first creating a set of possible fitness values for the given task (e.g., $x \in \{ 0, 1, 2, ... , 400 \}$), and then for each of these values ($x$) we calculate the ratio of solutions from the archive that have a fitness value less than or equal to $x$ in the case of CDF, or greater than $x$ in the case of CCDF. We perform this procedure on the archives returned from all independent evolutionary runs of $P_i$ (in order not to depend on a particular archive), as well as all evaluation scenarios, and record the median ratio for each possible fitness value. For example, if there are 20 evolutionary runs and 10 evaluation scenarios, this means that the median ratio is calculated over $20 \times 10 = 200$ numbers, for each possible fitness value.

By querying the CDF or CCDF for a certain fitness value that is considered ``good" in a given problem is akin to asking: what is the expected percentage of good solutions returned by $P_i$? For example, suppose that we are comparing $P_1$ and $P_2$ in our hexapod locomotion task (i.e., maximize walking speed). If we consider a walking speed of at least 0.3 m/s to be well-performing in our task, we can query the CCDF tables of $P_1$ and $P_2$ at the index that corresponds to the value of 0.3 m/s, and get their output values $F_{P_1}(0.3)$ and $F_{P_2}(0.3)$. If $F_{P_1}(0.3) = 0.4$, this means that \textit{randomly picking a solution from an archive returned by $P_1$ has a $0.4$ chance of being one with a walking speed of at least 0.3 m/s}. If $F_{P_2}(0.3) < F_{P_1}(0.3)$, this means that $P_2$ is not as effective as $P_1$ for the same performance level, thus, we can say that the quality of $P_2$ is lower than the quality of $P_1$ for a walking speed of at least 0.3 m/s in the considered scenario.

\section{Maze navigation experiments} \label{sec_maze_experiments}

\subsection{Experimental Setup}
\subsubsection{Simulation and Fitness Function}

\begin{figure}[t!]
    \centering
    \begin{subfigure}[t]{0.43\columnwidth}
        \centering
        \includegraphics[width=0.6\textwidth]{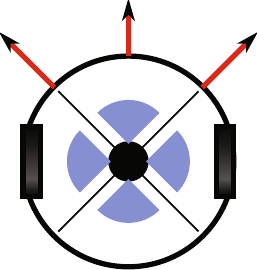}
        \caption{}
        \label{fig_maze_robot}
    \end{subfigure}%
    ~
    \begin{subfigure}[t]{0.43\columnwidth}
        \centering
        \includegraphics[width=0.6\textwidth]{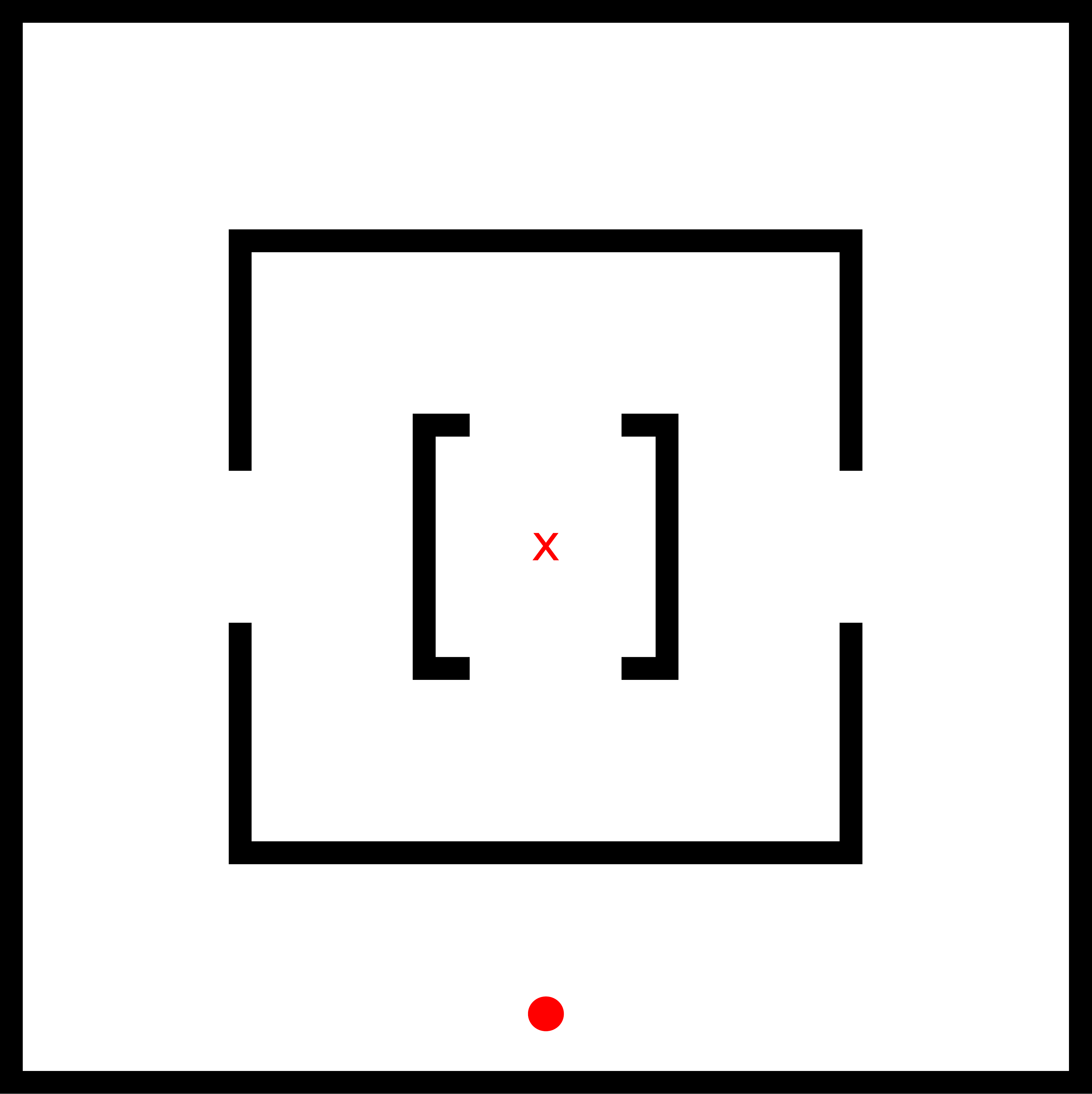}
        \caption{}
        \label{fig_maze}
    \end{subfigure}
    \caption{(a) Overview of the simulated mobile robot (diameter: 20 units). The robot is equiped with 3 laser range finder sensors (arrows) that return the normalized distance to the closest obstacle in that direction, and 4 pie-slice sensors that act as a compass towards the goal. The neural network controls the two motors that move the robot by setting their speed (a value between $[-2,2]$ units per simulation step). (b) The ``open" maze environment (size: 1000 $\times$ 1000 square units) employed in our experiments. The circle denotes the starting position and the cross denotes the goal.}
\end{figure}

We use a maze navigation task where a simulated mobile robot (Fig.~\ref{fig_maze_robot}) is controlled by an artificial neural network whose architecture and parameters are evolved (for parameter settings, see Appendix~\ref{sec_parameters}). The robot starts from the bottom of the arena and needs to reach the goal point at the center (Fig.~\ref{fig_maze}). At every simulation step, the Euclidean distance between the current position of the robot and the goal point is measured. The fitness function is the smallest distance achieved over the robot's lifetime~\cite{mouret2011novelty}, which is set to $3000$ simulation steps.

\subsubsection{Evaluation phase} \label{sec:mazeEvaluation}

We evaluate the archives of the EA-descriptor pairs in 16 different environments (shown in Fig.~\ref{fig_maze_best}). These environments are created by selectively blocking the openings of the ``open" maze environment to effectively allow only one realizable trajectory to the goal per environment.

\subsubsection{Behavioral Descriptors} \label{sec_maze_descriptors}

\begin{figure}[th]
	\centering
	\includegraphics[width=0.7\columnwidth]{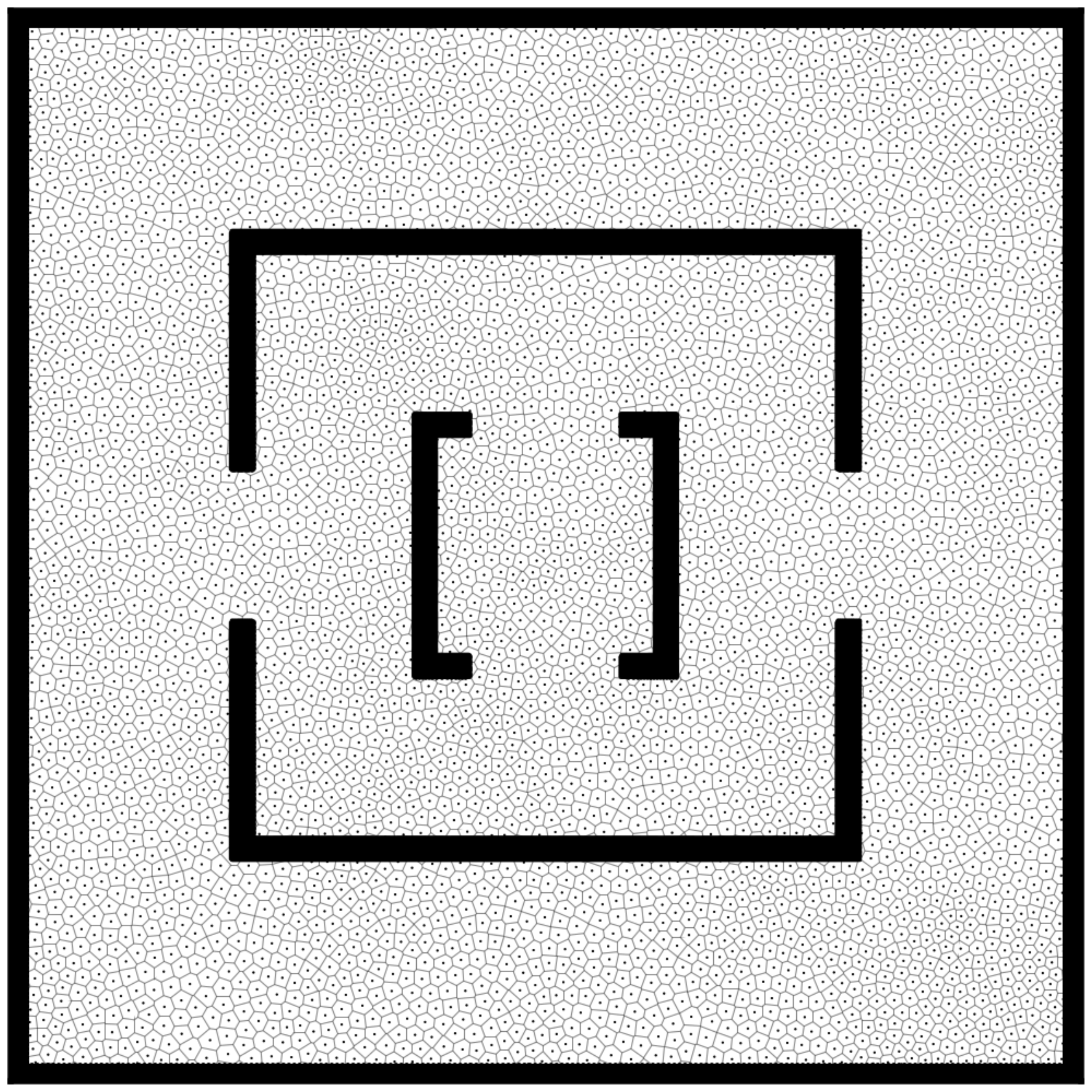}
	\caption{Environment used for the maze navigation task. The 5000 niches produced by the CVT routine using the 2D descriptor (i.e., that correspond to the possible end-locations of the mobile robot) are shown in the background.}
	\label{fig_maze_with_centroids}
\end{figure}

We use 6 behavioral descriptors of increasing dimensionality, that correspond to sampling (x,y) points along the trajectory of the robot. These are: 2D (end-point of the trajectory)\footnote{
See Fig.~\ref{fig_maze_with_centroids} for how the niches look like in 2D.
}, 10D (trajectory length = 5), 20D (trajectory length = 10), 50D (trajectory length = 25), 250D (trajectory length = 125), and 1000D (trajectory length = 500). MAP-Elites can only be used with the 2D, 10D and 20D descriptors, as more dimensions require more RAM than what is available. For example, in the 50D case, MAP-Elites requires 4096 TB of RAM just to store the matrix of pointers (not even the contents of the cells). For the 2D, 10D and 20D descriptors, we use a discretization of 71, 3 and 2 per dimension, resulting in 5041, 59049 and 1048576 cells, respectively.
For CVT-MAP-Elites, we set the number of niches $k=5000$ (see Appendix~\ref{sec_additionalExperiments} and Fig.~\ref{fig_maze_best_cluster_comparison}, \ref{fig_maze_cdf}A for a comparison between different values of $k$).

\subsubsection{Generating centroids}
The CVT algorithm relies on randomly sampled points that are clustered to find well-spread centroids. For the maze task, we cannot generate random trajectories by straightforwardly connecting random points in the arena because such trajectories would not match the physical constraints of the robot, which cannot teleport from any point to another in a single time step; and we cannot use a basic random walk because it would need too many samples to cover the space well. Instead, we generate a random point of the trajectory (e.g., the 42nd time step) within the bounds that are physically possible with the robot (here, 42 times $\pm 2$ units from the starting point, while staying within the bounds of the arena, that is, $1000 \times 1000$), then we generate another random point of the trajectory (e.g., the 23rd) with updated constraints (23 times $\pm 2$ units from the starting point, and 19 times $\pm 2$ units from the 42nd point), etc. We continue this process until we have chosen all the points of the trajectory.

\subsection{Results} \label{sec_maze_results}

\begin{figure*}[th]
	\centering
	\includegraphics[angle=0,width=\textwidth]{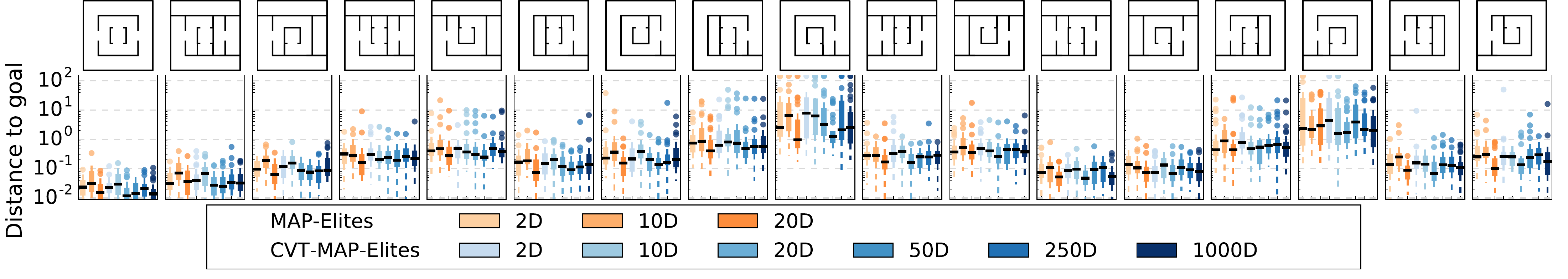}
	\caption{Best performance (distance to the goal, thus, lower is better) for each algorithm-descriptor pair in the ``open" maze environment (leftmost column) and all 16 evaluation environments which are created by selectively blocking certain paths that lead to the center of the maze (the goal). The box plots show the median (black line) and the interquartile range ($25^{th}$ and $75^{th}$ percentiles) over $30$ solutions; the whiskers extend to the most extreme data points not considered outliers, and outliers are plotted individually. In all cases, the median is below $10$ units, indicating that the algorithms have good overall performance with all behavioral descriptors in these $1000 \times 1000$ square unit environments. The behavior of CVT-MAP-Elites does not deteriorate in high-dimensional spaces (e.g., 1000D), illustrating that the algorithm can scale well. The $8^{th}$ and $14^{th}$ evaluation environments, where there are big outliers, are symmetrical and seem to be the most difficult ones, as they require an almost full clockwise turn followed by an almost full counterclockwise turn and vice versa.}
	\label{fig_maze_best}
\end{figure*}

\begin{figure*}[th]
	\centering
	\includegraphics[width=\textwidth]{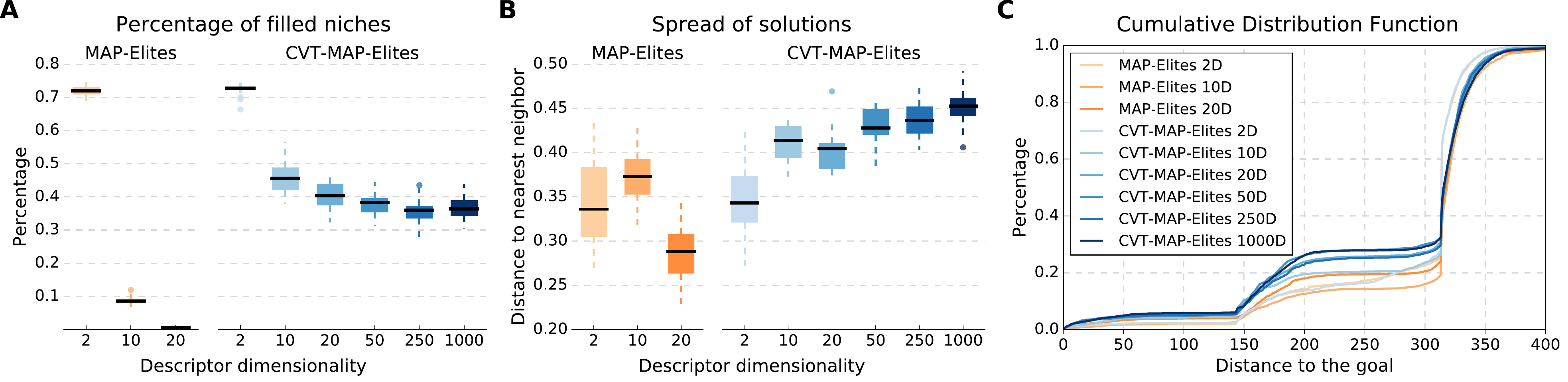}
	\caption{\textbf{(A)} Percentage of filled niches. \textbf{(B)} Spread of solutions measured as normalized distance to the nearest neighbor. The box plots show the median (black line) and the interquartile range ($25^{th}$ and $75^{th}$ percentiles) over $30$ solutions; the whiskers extend to the most extreme data points not considered outliers, and outliers are plotted individually. \textbf{(C)} The cumulative distribution function calculated as the median from all evaluation scenarios.}
	\label{fig_maze_results}
\end{figure*}

For all experiments, we use 30 independent evolutionary trials and 200k function evaluations. The results show that the expected best performance of both MAP-Elites with the 2D, 10D and 20D descriptors, and CVT-MAP-Elites with the additional 50D, 250D and 1000D descriptors remains constant over all the evaluation scenarios (Fig.~\ref{fig_maze_best}). The median distance is less than 2 units in most scenarios, while for the $8^{th}$ and $14^{th}$ environments, the median distance is still less than 10 units which is equal to the radius of the robot.

The median percentage of filled niches (Fig.~\ref{fig_maze_results}A) of MAP-Elites decreases sharply from 0.72 (2D) to 0.09 (10D) and 0.005 (20D). These values for CVT-MAP-Elites are 0.73 (2D), 0.46 (10D), 0.40 (20D), 0.38 (50D), 0.36 (250D) and 0.36 (1000D).
We also measure the spread of solutions ($s$) for each archive $\mathcal{A}$ using the following formula:
\begin{equation}
s = \frac{ \frac{1}{|\mathcal{A}|}\sum_{\mathbf{b} \in \mathcal{A}} d_{nn}(\mathbf{b},\mathcal{A})}{\max_{\mathbf{b} \in \mathcal{A}}d_{nn}(\mathbf{b},\mathcal{A})}
\end{equation}
where $\mathbf{b}$ is the behavioral descriptor of a solution in $\mathcal{A}$ and $d_{nn}(\mathbf{b},\mathcal{A})$ is the Euclidean distance of $\mathbf{b}$ to its nearest neighbor in $\mathcal{A}$. The median spread of solutions (Fig.~\ref{fig_maze_results}B) for MAP-Elites increases from 0.34 (2D) to 0.37 (10D) and then decreases to 0.29 (20D). Interestingly, in CVT-MAP-Elites the solutions become more well-spread in higher dimensions: 0.34 (2D), 0.41 (10D), 0.40 (20D), 0.43 (50D), 0.44 (250D) and 0.45 (1000D).

The expected quality of the archives (Fig.~\ref{fig_maze_results}C) for a distance of at most 20 units (diameter of the robot) is slightly higher with CVT-MAP-Elites (MAP-Elites vs CVT-MAP-Elites): 2D: 0.02 vs 0.02; 10D: 0.01 vs 0.02; 20D: 0.02 vs 0.03; for the additional descriptors of CVT-MAP-Elites these values are: 0.04 (50D) and 0.03 (250D, 1000D). Overall, these results indicate that CVT-MAP-Elites performs as well as MAP-Elites, but can scale to high dimensions.

\section{Hexapod locomotion experiments}

\subsection{Experimental Setup}

\subsubsection{Simulation and Fitness Function}

We use the hexapod locomotion task of~\cite{cully2015robots} using the Dynamic Animation and Robotics Toolkit\footnote{DART can be found in https://github.com/dartsim/dart}. The controller is designed to be a simple, open-loop oscillator that actuates each servo by a periodic signal of frequency 1Hz, parameterized by 3 values: the amplitude of oscillation, its phase shift and its duty cycle (i.e., the fraction of each period that the joint angle is positive). Each leg has 3 joints, however, only the movement of the first 2 is defined in the parameters\footnote{The control signal of the third servo of each leg is the opposite of the second one.}~\cite{cully2015robots}. Therefore, there are 6 parameters per leg, thus, 36 parameters for controlling the whole robot. The fitness function is the forward distance covered in 5 seconds (thus, we maximize walking speed).

\subsubsection{Evaluation phase} \label{sec:hexapodEvaluation}
\begin{figure*}[th!]
	\centering
	\includegraphics[angle=0,width=\textwidth]{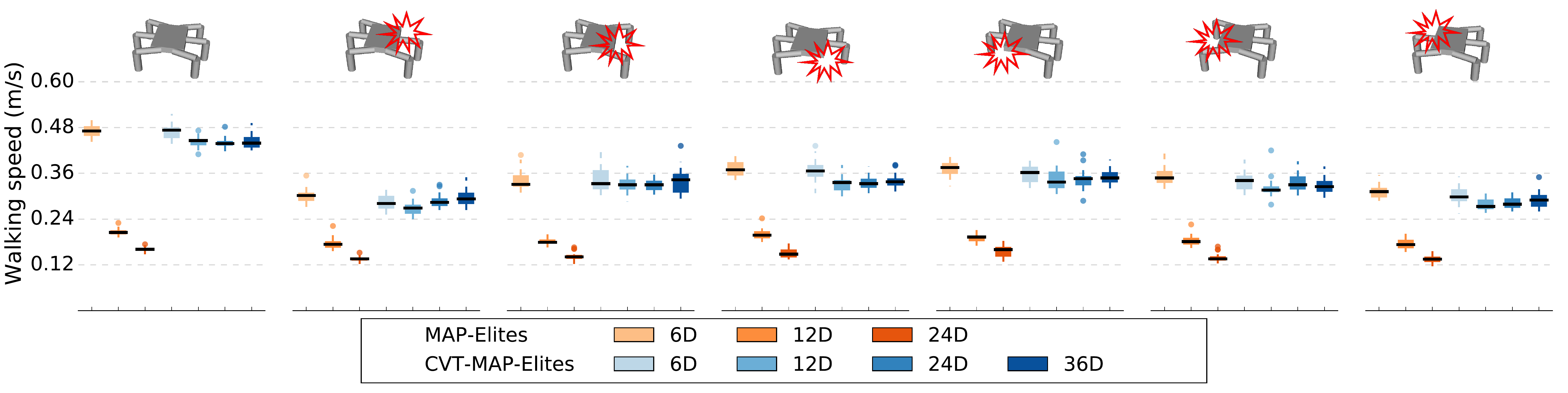}
	\caption{Best performance (walking speed, thus, higher is better) for each algorithm-descriptor pair in the undamaged case and the $6$ damage conditions which correspond to removing a different leg of the hexapod robot. The box plots show the median (black line) and the interquartile range ($25^{th}$ and $75^{th}$ percentiles) over $20$ solutions; the whiskers extend to the most extreme data points not considered outliers, and outliers are plotted individually. The performance of the best solutions found by MAP-Elites diminishes as the dimensionality of the descriptor increases. In contrast, CVT-MAP-Elites retains its performance with the increased dimensionality, thus, it scales significantly better than MAP-Elites.}
	\label{fig_best_performance}
\end{figure*}

During the evolutionary phase, we use a model of an intact robot to generate the archives, whereas, during the evaluation phase, we evaluate 6 damage cases that correspond to removing a different leg from the hexapod robot (see Fig.~\ref{fig_best_performance}).

\subsubsection{Behavioral Descriptors} \label{sec_hexapod_descriptors}

We use 4 behavioral descriptors of increasing dimensionality.
For all experiments, CVT-MAP-Elites uses $k=10$k niches (see Appendix~\ref{sec_additionalExperiments} and Fig.~\ref{fig_hexapod_best_cluster_comparison}, \ref{fig_hexapod_ccdf}A for a comparison between different values of $k$).
%
\begin{itemize}
  \item Duty Factor (6D): It is defined as the proportion of time that each leg is in contact with the ground:
  \begin{equation}
  \mathbf{b} = \left [\frac{\sum_{t=1}^{T}C_{1}(t)}{T}, \: ... \: ,\frac{\sum_{t=1}^{T}C_{6}(t)}{T} \right ] \in \mathbb{R}^6
  \end{equation}
  where $\mathbf{b}$ is the descriptor, $C_i(t)$ denotes the Boolean value of whether leg $i$ is in contact with the ground at time $t$ (i.e., 1: contact, 0: no contact), recorded at each time step (every 15 ms) and averaged over the number of time steps $T$ of the simulation. For MAP-Elites, we use 5 discretizations per dimension, resulting in 15625 bins.

  \item 12D subset of controller parameters that contains the phase shift for each of the 12 joints. For MAP-Elites, we use 3 discretizations per dimension (thus, $3^{12}$ bins).

  \item 24D subset of controller parameters that contains the amplitude of oscillation and the phase shift for each of the 12 joints. For MAP-Elites, we use 2 discretizations per dimension (resulting in nearly 17 million bins).

  \item Controller Parameters (36D): In this case, the behavior space is the same as the parameter space.
  MAP-Elites cannot be employed since even using 2 discretizations per dimension requires more than 256 GB of RAM.
\end{itemize}

\subsection{Results} \label{sec_hexapod_results}

\begin{figure*}[th]
	\centering
	\includegraphics[width=\textwidth]{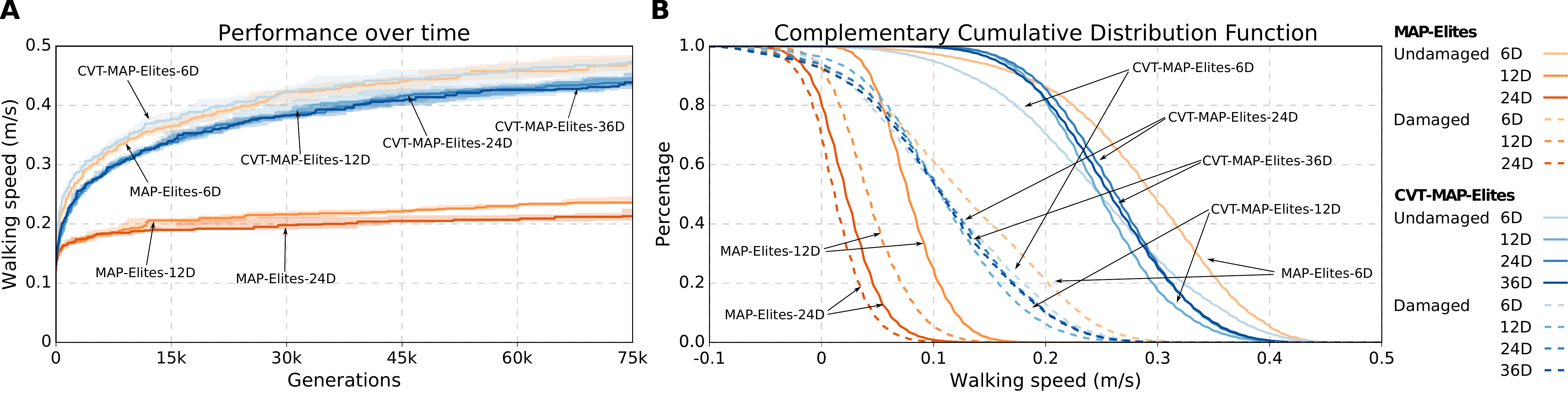}
	\caption{\textbf{(A)} Median performance of the best solution found over the generations. The light zones represent the $25^{th}$ and $75^{th}$ percentiles. \textbf{(B)} The complementary cumulative distribution function for both the undamaged case (shown with \textit{solid} lines) and all damage conditions (aggregated and shown with \textit{dotted} lines). As the dimensionality of the descriptor increases, the performance of the solutions returned by MAP-Elites deteriorates. In contrast, CVT-MAP-Elites maintains the same level of performance, thus, scaling significantly better than MAP-Elites. Solutions with a negative walking speed signify that the hexapod robot moves backward.}
	\label{fig_fitness_aggregate}
\end{figure*}

For all experiments, we use 20 independent evolutionary runs\footnote{
%
We perform only 20 runs as each required more than 24 hours of computation time on modern (2015) 12-core Intel Xeon CPUs.
} and 75k generations. Overall, our results indicate that during the evolutionary phase (Fig.~\ref{fig_fitness_aggregate}A), the median performance of the best individuals of CVT-MAP-Elites is approximately the same when using the 12D, 24D and 36D descriptors, despite the increase in dimensionality. When using the 6D descriptor, the performance is slightly better, however, this is likely due to the fact that this descriptor is calculated in behavior space, whereas the others in parameter space~\cite{mouret2012encouraging, doncieux2014beyond}. With MAP-Elites, the performance of the best individuals when using the 6D descriptor is approximately the same as in CVT-MAP-Elites. However, when the dimensionality of the descriptor increases, the performance of MAP-Elites deteriorates significantly.

In both the undamaged and damaged conditions, the expected best performance of MAP-Elites decreases with each increase in descriptor dimensionality, whereas the one of CVT-MAP-Elites remains relatively constant (MAP-Elites vs CVT-MAP-Elites; median gaits given in m/s): undamaged: [6D: 0.47 vs 0.47; 12D: 0.21 vs 0.45; 24D: 0.16 vs 0.44; 36D (CVT-MAP-Elites): 0.44]; damaged (average over 6 settings): [6D: 0.34 vs 0.33; 12D: 0.18 vs 0.31; 24D: 0.15 vs 0.32; 36D (CVT-MAP-Elites): 0.32].

The results of both the undamaged and damaged conditions indicate that MAP-Elites-6D has higher probability of finding better solutions than all other EA-descriptor pairs (Fig.~\ref{fig_fitness_aggregate}B). CVT-MAP-Elites-6D does not perform as well, most likely due to the fact that it uses 10k niches during evolution, whereas MAP-Elites-6D uses 1.5 times more ($5^6$); this indicates that the number of niches for CVT-MAP-Elites could be better tuned, though this is beyond the scope of this study.

The expected quality of the archives produced by MAP-Elites significantly decreases with higher-dimensional descriptors, whereas with CVT-MAP-Elites it is not (Fig.~\ref{fig_fitness_aggregate}B). For all the damaged cases and in particular for a walking speed of 0.2 m/s, the expected percentage of solutions returned by MAP-Elites that have at least this value is 21.3\% for 6D, 0\% for 12D and 0\% for 24D; for CVT-MAP-Elites these are 13.3\% for 6D, 6.5\% for 12D, 10.8\% for 24D and 10.5\% for 36D. Comparing the two algorithms using the 12D and 24D descriptors reveals that the differences are highly significant ($p<10^{-44}$, Mann-Whitney U test). Thus, randomly choosing among the solutions returned by both algorithms, we obtain higher quality ones with CVT-MAP-Elites, as the descriptor dimensionality increases.

\section{Discussion and Conclusion}

We have shown that CVT-MAP-Elites can be applied in problems where the dimensionality is prohibitive for MAP-Elites (e.g., 1000 dimensions in the maze experiments). We have additionally demonstrated that in the hexapod locomotion tasks, CVT-MAP-Elites found gaits that were on average 1.7 to 2.1 times faster (during the evaluation settings) with the corresponding increase in feature space dimensionality.
This is because CVT-MAP-Elites has a more precise control over the balance between diversity and performance. For example, there is more selective pressure for performance when randomly selecting a parent from an archive of 1000 elites than from an archive of 1 million because the niches are bigger in the former case than in the latter (an elite from a big niche ``reign'' on more solutions). In the extreme case of a single niche, MAP-Elites and CVT-MAP-Elites would act like a stochastic hill climber, that is, with a very strong pressure for performance. In MAP-Elites the increase in dimensionality exponentially increases the number of niches, and as these niches get filled with solutions, selective pressure for performance decreases. In CVT-MAP-Elites, by having fewer niches (thus, solutions) and keeping the same selection method (e.g., 100 parents uniformly at random at every generation), we effectively increase selective pressure for performance.

In all our experiments, we chose to perform distance calculations using the Euclidean norm because it is the distance function used in other quality diversity studies \cite{cully2017,pugh2015confronting,lehman2011evolving}; however, ``fractional'' norms (Minkowski distance of order $p<1$) might be more appropriate for high-dimensional spaces, because they provide a better contrast between the farthest and the nearest neighbor~\cite{aggarwal2001surprising, xia2015effectiveness}. Nevertheless, additional experiments revealed that results are not qualitatively affected by a change to fractional norms, and, in particular, that there is no significant difference between a Euclidean and a fractional norm when considering the generalization performance (see Fig. \ref{fig_maze_best_distance_comparison}, \ref{fig_maze_cdf}B, \ref{fig_hexapod_best_distance_comparison}, \ref{fig_hexapod_ccdf}B).

CVT-MAP-Elites is a natural extension of MAP-Elites since it behaves like the latter in low-dimensional spaces, if given the same amount of well-spread niches. In addition, it does not require any major modifications over MAP-Elites. This is because the CVT routine (not part of the main EA) is responsible for generating the centroids and only needs to run once, before the EA starts; thus, the EA only needs to load these centroids. One can easily substitute the CVT routine with their own implementation, without any change in the EA. Interestingly, such a routine could be designed in a way that places more centers (thus, more variation) along certain dimensions. To ease deployment, we provide a python script for generating the centroids (see Appendix~\ref{sec_sourcecode}).

The computational complexity of the method we provide for constructing the CVT (in Alg.~\ref{algo:cvt}) is $O(ndki)$, where $n$ is the number of $d$-dimensional samples to be clustered, $k$ is the number of clusters and $i$ is the number of iterations needed until convergence. From our experiments, we noticed that the clustering does not need to be very precise when using such a large number of clusters (i.e., thousands), thus, the number of iterations $i$ can be fixed to limit the overall complexity of CVT-MAP-Elites. Nevertheless, one could resort to other CVT construction methods, which could offer significant speed-ups~\cite{hateley2015fast}.
%
Furthermore, finding the centroid closest to a given behavior descriptor can be accelerated using data structures such as \textit{k}-d trees~\cite{bentley1975multidimensional} or others that exploit the characteristics of the Voronoi tessellation (e.g., see~\cite{sharifzadeh2010vor},~\cite{aurenhammer2000voronoi}).

A key factor that enabled CVT-MAP-Elites to scale to 1000 dimensions in the maze experiments is the proper sampling of trajectories to generate the CVT and, as a consequence, the generation of centroids that more closely approximate trajectories that could be followed by the robot. A naive approach for sampling trajectories would sample each element along the trajectory from $[0,1000]^2$ in the $1000 \times 1000$ arena. This, however, would result in unrealistic centroid trajectories, because the robot cannot move in a single time step to any point of the arena (it is constrained by its maximum speed). The behavior descriptors of the hexapod experiments do not have this problem, as they are properly defined in $[0,1]^d$ (i.e., the behavior space is \emph{dense}). Nevertheless, care needs to be taken when sampling the behavior space of interest during CVT construction: the centroids extracted from the samples have to be representatives of potential behaviors that make sense in the task.

While it is easy to sample behavior descriptors in many evolutionary robotics tasks, there exist many others for which sampling realistic behavior descriptors might prove challenging, thus, complicating the use of CVT-MAP-Elites. This is typically the case for behavior descriptors that involve trajectories of dynamical systems in high-dimensional state spaces. To put it clearly, CVT-MAP-Elites allows MAP-Elites to scale up to high-dimensional spaces, but it is still a \emph{grid-based} algorithm~\cite{cully2017}, which is a category of quality diversity algorithms that is especially effective when the behavior space can be divided beforehand~\cite{cully2017}. By contrast, if the bounds of the space are unknown before evolution, or if the set of possible behavior descriptors is heavily constrained\footnote{Typically, when the behavior space is \emph{sparse}, i.e., many behavior descriptor combinations are impossible to obtain in the considered system.}, then vanilla CVT-MAP-Elites is probably not the most appropriate algorithm.

This issue can be potentially mitigated by letting the solutions generated by the evolutionary algorithm define the bounds of the space of interest. Thus, instead of pre-calculating the CVT, we can periodically recalculate it throughout the evolutionary run based on whether the bounds have changed~\cite{vassiliades2017unbounded}. In addition, it is possible to change the bounding volume to a different shape than a hyper-rectangle, as well as to vary the density of the niches so that to have a greater number on the outer part of the volume: such a change could potentially put more pressure towards expanding the volume and explore novel regions in behavior space. At any rate, other quality diversity algorithms could be more effective when nothing is known about the behavior space beforehand \cite{cully2017}. In these cases, it is worth investigating algorithms that rely on distance computations between different generated points~\cite{vassiliades2017multimodal} (such as Novelty Search with Local Competition~\cite{lehman2011evolving} or Restricted Tournament Selection~\cite{harik1995rts}), rather than between points and fixed centroids.

\appendices

\section{Additional Experiments} \label{sec_additionalExperiments}

In this section, we report additional experiments where we vary the number of niches $k$ and the distance metrics used both when generating the centroids (during the CVT construction) and when finding the closest centroid (throughout the evolutionary run).

\subsubsection*{Number of niches}
In the maze navigation task, we vary $k \in \{5,50,500,5000\}$ and show results with 3 different descriptors (2D, 20D, 250D). In the hexapod locomotion task, we vary $k \in \{10,100,1000,10000,100000\}$ using the duty factor (6D) behavioral descriptor.

\subsubsection*{Distance metrics}
We use the following formula which defines the Minkowski distance of order $p$ between two points $\mathbf{x} \in \mathbb{R}^d$ and $\mathbf{y} \in \mathbb{R}^d$:
\begin{equation}
  ||\mathbf{x}-\mathbf{y}||_{p} = \Big(\sum_{i=1}^{d}|\mathbf{x}_i - \mathbf{y}_i|^p \Big)^{1/p}
\end{equation}
By setting $p=2$, we obtain the Euclidean distance metric. As shown in~\cite{aggarwal2001surprising}, as $p$ increases, the contrast between the farthest and nearest neighbor becomes poorer, especially in higher dimensional spaces. By setting $p<1$, we obtain a ``fractional distance metric''~\cite{aggarwal2001surprising} which has been shown to be better suitable in high dimensional spaces~\cite{aggarwal2001surprising}, despite not being a true metric (as it violates the triangle inequality). In both tasks, we compare the Euclidean distance metric ($p=2$) with a fractional one of order $p=0.5$. More specifically, in the maze navigation task, we perform such a comparison using 3 different descriptors (50D, 250D, 1000D); in the hexapod locomotion task, we use the full controller space (36D) as the behavior space.

\subsubsection*{Maze Navigation Results}

\begin{figure*}[th]
	\centering
	\includegraphics[width=\textwidth]{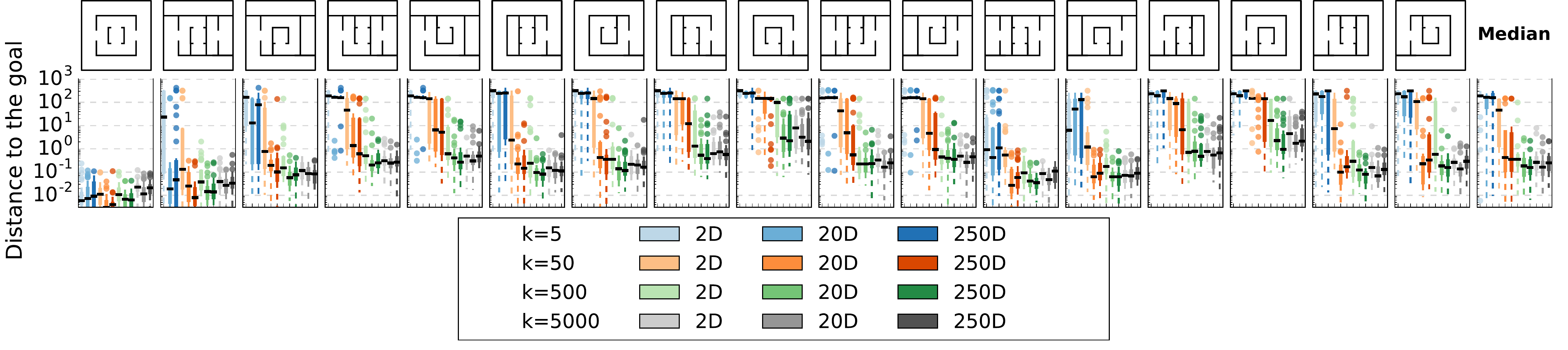}
	\caption{Comparison of different values of cluster parameter $k$ (number of niches/centroids) for the expected best performance (distance to the goal) in the ``open'' maze environment (leftmost column) and all 16 evaluation environments each of which permits a single path to the goal (center). The box plots show the median and the interquartile range over 30 solutions, apart from the rightmost column which is calculated from the medians over all 17 environments. As $k$ increases, the expected best performance increases as well, however, there is no significant difference between $k=500$ and $k=5000$.}
	\label{fig_maze_best_cluster_comparison}
\end{figure*}

\begin{figure*}[th]
	\centering
	\includegraphics[width=\textwidth]{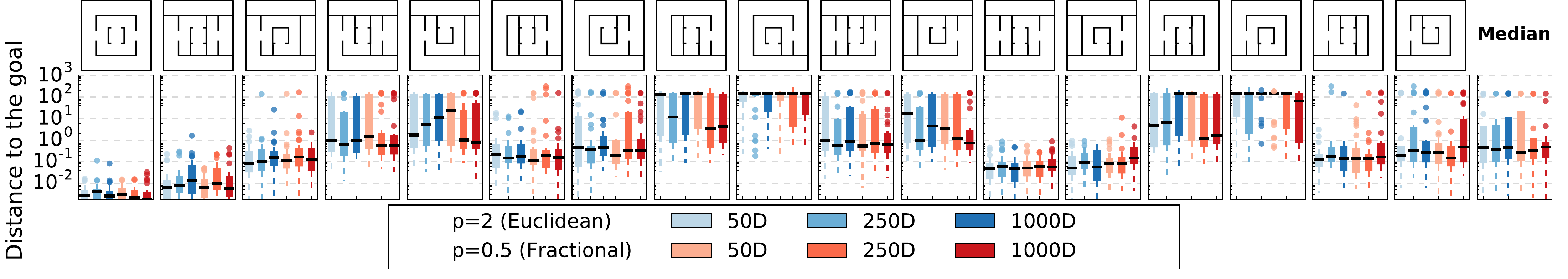}
	\caption{Comparison of distance metrics for the expected best performance (distance to the goal) in the ``open'' maze environment (leftmost column) and all 16 evaluation environments each of which permits a single path to the goal (center). The box plots show the median and the interquartile range over 30 solutions, apart from the rightmost column which is calculated from the medians over all 17 environments. There is no significant difference between the two distance metrics.}
	\label{fig_maze_best_distance_comparison}
\end{figure*}

\begin{figure*}[th]
	\centering
	\includegraphics[width=\textwidth]{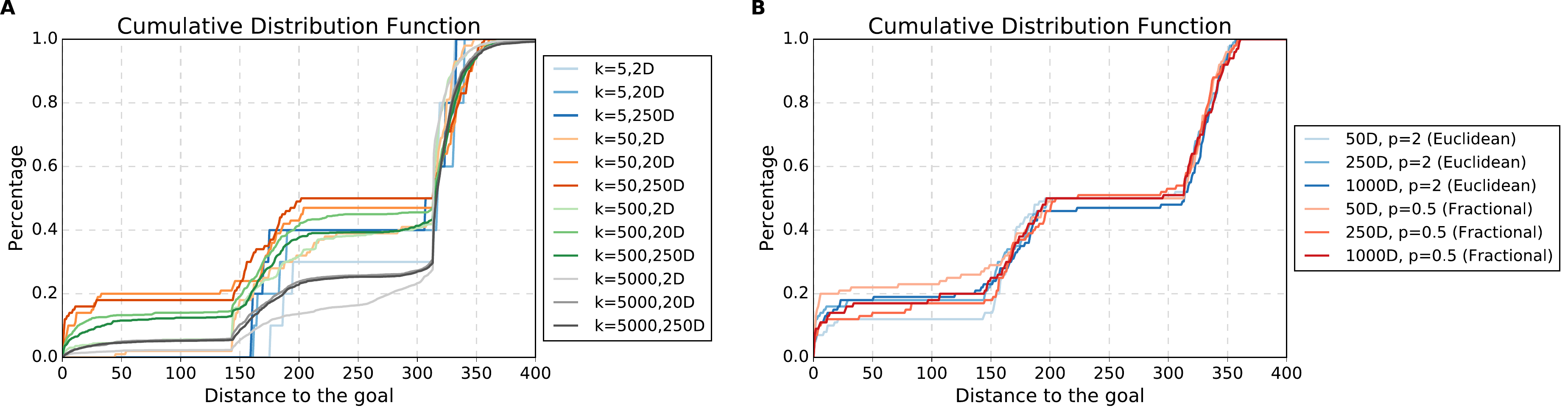}
	\caption{Cumulative Distribution Function for the maze navigation experiments. \textbf{(A)} The percentage of good solutions (i.e., distance less than 100) is higher when using the higher dimensional descriptors (20D, 250D) and when using $k=50$. \textbf{(B)} There is no significant difference between the two distance metrics.}
	\label{fig_maze_cdf}
\end{figure*}

In the maze navigation task, the expected best performance increases with the increase of $k$ (Fig.~\ref{fig_maze_best_cluster_comparison}). The worst performance is achieved with $k=5$, however, this is expected as it is equivalent to using a population size of at most 5. Parameter $k$ can be set according to the user's preferences: for example, it makes sense to have a high value when visualizing search spaces~\cite{mouret2015illuminating}. The cumulative distribution function (CDF) in Fig.~\ref{fig_maze_cdf}A shows that when using a $k=50$, the percentage of ``good'' solutions (i.e., where the distance to the goal is less than 100) is higher than when using $k=500$ or $k=5000$. This is expected since having more niches allows the algorithm to retain more locally optimal solutions. This also shows that CVT-MAP-Elites can work well in this problem even with a population size of at most 50. Fig.~\ref{fig_maze_cdf}A additionally shows that the expected quality of the archives is higher when using the higher dimensional descriptors (i.e., 20D or 250D); this is expected since the 2D descriptor cannot capture multiple trajectories reaching the same 2D location.
There is no significant difference between the two distance metrics as shown in Fig.~\ref{fig_maze_best_distance_comparison} and Fig.~\ref{fig_maze_cdf}B.

\subsubsection*{Hexapod Locomotion Results}

\begin{figure*}[th]
	\centering
	\includegraphics[width=\textwidth]{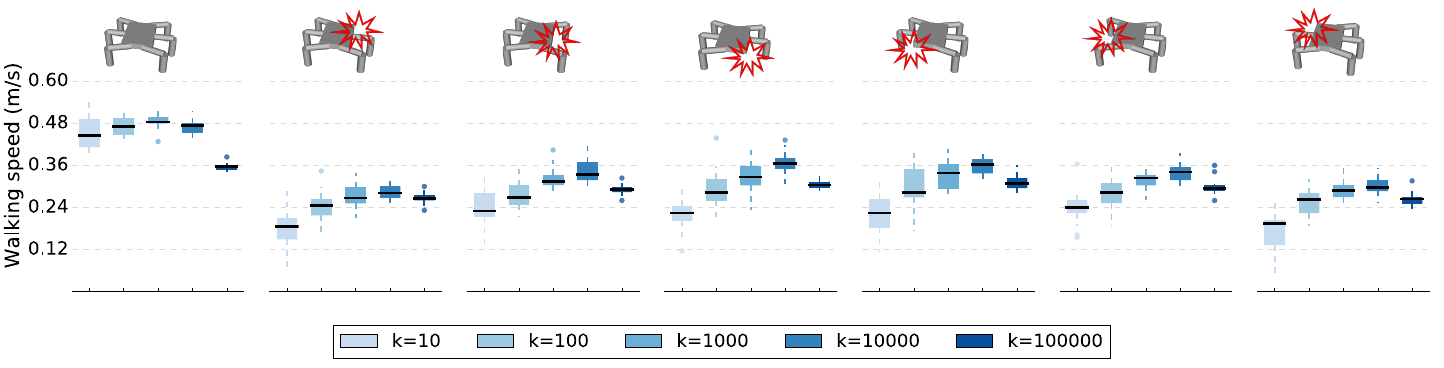}
	\caption{Comparison of different values of cluster parameter $k$ (number of niches/centroids) for the expected best performance (walking speed) in the hexapod locomotion task of the evolutionary setting (undamaged robot, leftmost column) and the evaluation settings (damaged robot). The box plots show the median and the interquartile range over 20 independent solutions (after 75k generations).}
	\label{fig_hexapod_best_cluster_comparison}
\end{figure*}

\begin{figure*}[th]
	\centering
	\includegraphics[width=\textwidth]{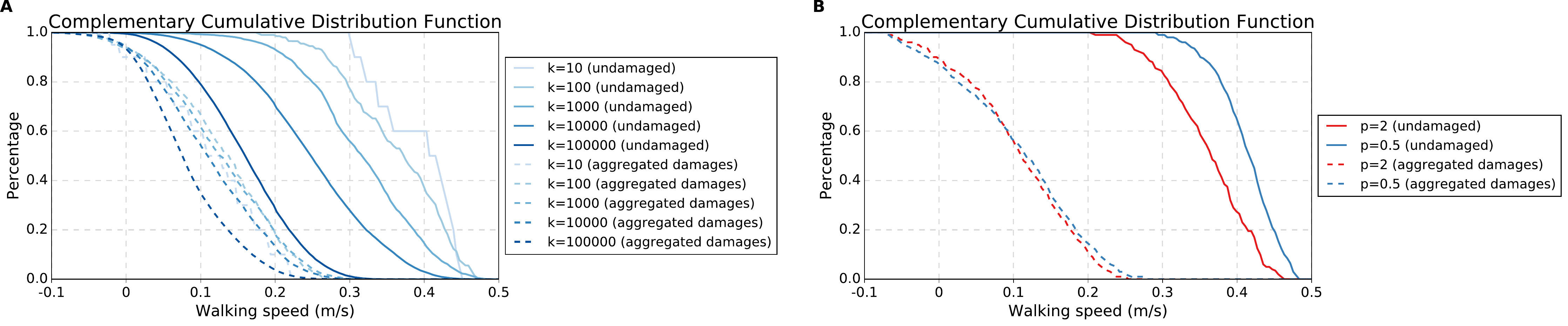}
	\caption{Complementary Cumulative Distribution Function for the hexapod locomotion experiments. \textbf{(A)} In the undamaged case, a smaller $k$ is preferable. However, when testing for generalization (damaged cases) there is no significant difference when $k \in \{10,100,1000,10000\}$. \textbf{(B)} The fractional norm is preferable when considering only the evolutionary performance (undamaged case), however, there is no significant difference between the two norms when considering the generalization performance (median over damaged cases).}
	\label{fig_hexapod_ccdf}
\end{figure*}

In the hexapod locomotion task, the expected best performance is not significantly affected when the robot is undamaged and when $k \in \{10,100,1000,10000\}$; however, when $k=100000$ there is a significant decrease (Fig.~\ref{fig_hexapod_best_cluster_comparison}). In the damaged cases, the best performance is achieved with $k=10000$, while $k=10$ has the worst results. The complementary CDF (CCDF) shown in Fig.~\ref{fig_hexapod_ccdf}A shows that it is more preferable to use smaller $k$ if we care about the evolutionary performance (undamaged robot). However, when we care about the generalization performance (evaluation settings - damaged cases), there is no significant difference when $k \in \{10,100,1000,10000\}$, while $k=100000$ has the worst expected performance.

\begin{figure*}[th]
	\centering
	\includegraphics[width=\textwidth]{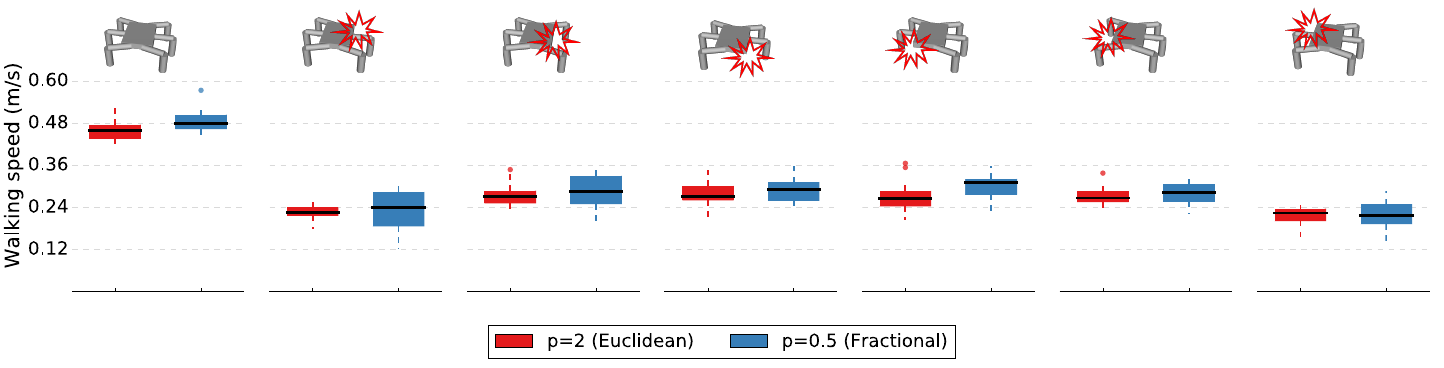}
	\caption{Comparison of distance metrics for the expected best performance (walking speed) in the hexapod locomotion task of the evolutionary setting (undamaged robot, leftmost column) and the evaluation settings (damaged robot). The box plots show the median and the interquartile range over 20 independent solutions (after 25k generations).}
	\label{fig_hexapod_best_distance_comparison}
\end{figure*}

The expected best performance in the undamaged case is slightly greater when using the fractional norm (Fig.~\ref{fig_hexapod_best_distance_comparison}). This is also apparent from the CCDF in Fig.~\ref{fig_hexapod_ccdf}B illustrating that the fractional norm has advantages when we care about the evolutionary performance. In the damaged cases, however, there is no significant difference between the performance values of the two norms in both the expected best performance (Fig.~\ref{fig_hexapod_best_distance_comparison}) and the expected quality of the solutions (Fig.~\ref{fig_hexapod_ccdf}B).

\section{Source Code} \label{sec_sourcecode}
The source code of the experiments can be found in {\footnotesize\url{https://github.com/resibots/vassiliades_2017_cvt_map_elites}}.

\section{Parameters of Evolutionary Algorithms} \label{sec_parameters}

The following table summarizes the parameters used for the evolutionary algorithms in the experiments reported in the main text.

\begin{table}[H]
	\begin{tabular}{ c c c }
		\toprule
		\textsc{\textbf{Common EA}} & \textsc{\textbf{Maze}} & \textsc{\textbf{Hexapod}} \\
		\textsc{\textbf{Parameters}} & \textsc{\textbf{Experiments}} & \textsc{\textbf{Experiments}} \\
		\midrule
		initial \#neurons / \#connec. & $\in [0,20] / [0,40]$ & N/A \\
		connec. add./del./modif. rate & 0.15 & N/A \\
		neuron (tanh) add./del. rate & 0.1 & N/A \\
		\#parameters in controller & variable & 36\\
		parameter values & $\{-2,-1.5,...,2 \}$ & $\{0,0.05,...,1 \}$\\
		initial \#solutions & 2k & 10k \\
		\#offspring per gen. & 200 & 200 \\
		\#gen. (eval.) & 990 (200k) & 75k (15010200) \\
		\midrule
		\textsc{\textbf{MAP-Elites}} & & \\
		\midrule
		behav. dimensions &  \multirow{2}{*}{2(71),10(3),20(2)} &  \multirow{2}{*}{6(5),12(3),24(2)} \\
		(\#discret. per dim.) & & \\
		\midrule
		\textsc{\textbf{CVT-MAP-Elites}} & & \\
		\midrule
		behav. dimensions & 2,10,20,50,250,1000 & 6,12,24,36 \\
		\#niches ($k$) & 5k & 10k\\
		\#samples for CVT ($K$) & 1000k & 100k\\
		\#iterations for kmeans & until convergence & until convergence \\
		\bottomrule
	\end{tabular}
  \label{tab_parameters}
\end{table}

\ifCLASSOPTIONcaptionsoff
  \newpage
\fi



\bibliographystyle{IEEEtran}
\bibliography{IEEEabrv,./bib/refs}

%





\end{document}